\documentclass[runningheads]{llncs}

\usepackage{subcaption}
\usepackage{bbding}
\usepackage{pifont}
\usepackage{utfsym}
\usepackage{eccv}

\usepackage{eccvabbrv}
\usepackage{makecell}
\usepackage{amssymb,mathrsfs,amsmath}
\usepackage{multirow}
\usepackage{array}
\usepackage{setspace}
\usepackage{tabularx}
\usepackage{graphicx}
\usepackage{booktabs}

\usepackage[accsupp]{axessibility}  

\usepackage{hyperref}

\usepackage{orcidlink}

\begin{document}

\title{VSViG: Real-time Video-based Seizure Detection via Skeleton-based Spatiotemporal ViG} 

\titlerunning{VSViG: Video-based Seizure Detection via Skeleton-based ST-ViG}

\author{Yankun Xu\inst{1} \and
Junzhe Wang\inst{1} \and
Yun-Hsuan Chen\inst{1}\and
Jie Yang\inst{1}\and
Wenjie Ming\inst{2}\and
Shuang Wang\inst{2}\and
Mohamad Sawan\inst{1,*}}

\authorrunning{Y. Xu, et al.}

\institute{CenBRAIN Neurotech, Westlake University, Hangzhou, China 
\email{\{xuyankun,wangjunzhe,chenyunxuan,yangjie,sawan\}@westlake.edu.cn} \and
Epilepsy Center, SAHZU, Zhejiang University, Hangzhou, China\\
\email{\{hflmwj,wangs77\}@zju.edu.cn}}

\maketitle

\begin{abstract}
An accurate and efficient epileptic seizure onset detection can significantly benefit patients. Traditional diagnostic methods, primarily relying on electroencephalograms (EEGs), often result in cumbersome and non-portable solutions, making continuous patient monitoring challenging. The video-based seizure detection system is expected to free patients from the constraints of scalp or implanted EEG devices and enable remote monitoring in residential settings. 
Previous video-based methods neither enable all-day monitoring nor provide short detection latency due to insufficient resources and ineffective patient action recognition techniques. Additionally, skeleton-based action recognition approaches remain limitations in identifying subtle seizure-related actions. To address these challenges, we propose a novel Video-based Seizure detection model via a skeleton-based spatiotemporal Vision Graph neural network (VSViG) for its efficient, accurate and timely purpose in real-time scenarios. Our experimental results indicate VSViG outperforms previous state-of-the-art action recognition models on our collected patients' video data with higher accuracy (5.9\% error), lower FLOPs (0.4G), and smaller model size (1.4M). Furthermore, by integrating a decision-making rule that combines output probabilities and an accumulative function, we achieve a 5.1 s detection latency after EEG onset, a 13.1 s detection advance before clinical onset, and a zero false detection rate. The project homepage is available at: \url{https://github.com/xuyankun/VSViG/}
\end{abstract}

\section{Introduction}
\label{sec:Intro}
As a common neurodegenerative disorder, epilepsy affects approximately $1\%$ population worldwide \cite{moshe2015epilepsy,thijs2019epilepsy}. A reliable epileptic seizure onset detection method can benefit patients significantly, because such a system equipped with accurate algorithms can promptly alert a seizure onset. According to previous studies \cite{tang2021self,shoeb2010application,hussein2018epileptic}, they predominantly focused on designing seizure detection algorithms based on electroencephalograms (EEGs). Although EEG can sense subtle brain changes just after a seizure begins, the use of scalp or implantable EEG devices often causes discomfort to patients and restricts them to hospital epilepsy monitoring units (EMUs). Consequently, there is a growing interest in developing an accurate video-based seizure detection system that could alleviate the discomfort associated with EEG helmets and facilitate remote monitoring of epileptic patients in residential settings \cite{videomonitoring}. However, the development of video-based seizure detection is hindered by several challenges as follows:

\textbf{(1) Lack of datasets.} 
Seizure-related video data collection is substantially time-consuming and requires doctoral expertise to annotate the different seizure-related periods. And, surveillance video recordings from the hospital EMUs contain highly sensitive information. Based on these reasons, there is no public video data intended for seizure study yet. Although \cite{perez2021transfer} released a related dataset, they only implement action anticipation on ictal videos which cannot provide any false detection rate (FDR) information.

\begin{figure}[t]
  \centering

  \includegraphics[width=0.75\columnwidth]{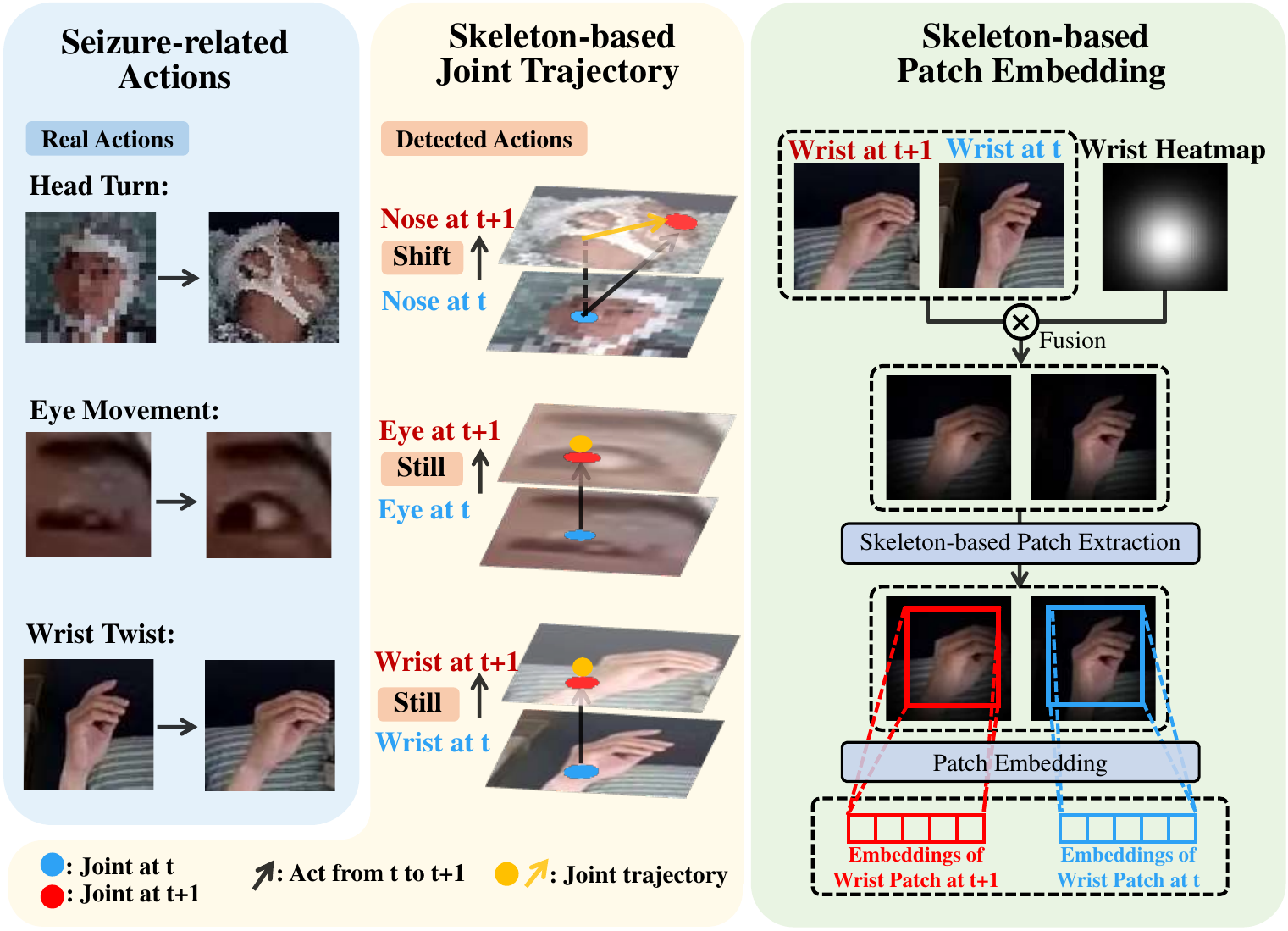}
  \caption{Motivation of proposed skeleton-based patch embedding. The left shows real seizure-related actions; The middle shows challenges of traditional skeleton-based approaches; The right shows our strategy to address challenges.}
  \label{fig:motivate}
\end{figure}

\begin{figure}[!t]
  \centering
    \includegraphics[width=0.6\columnwidth]{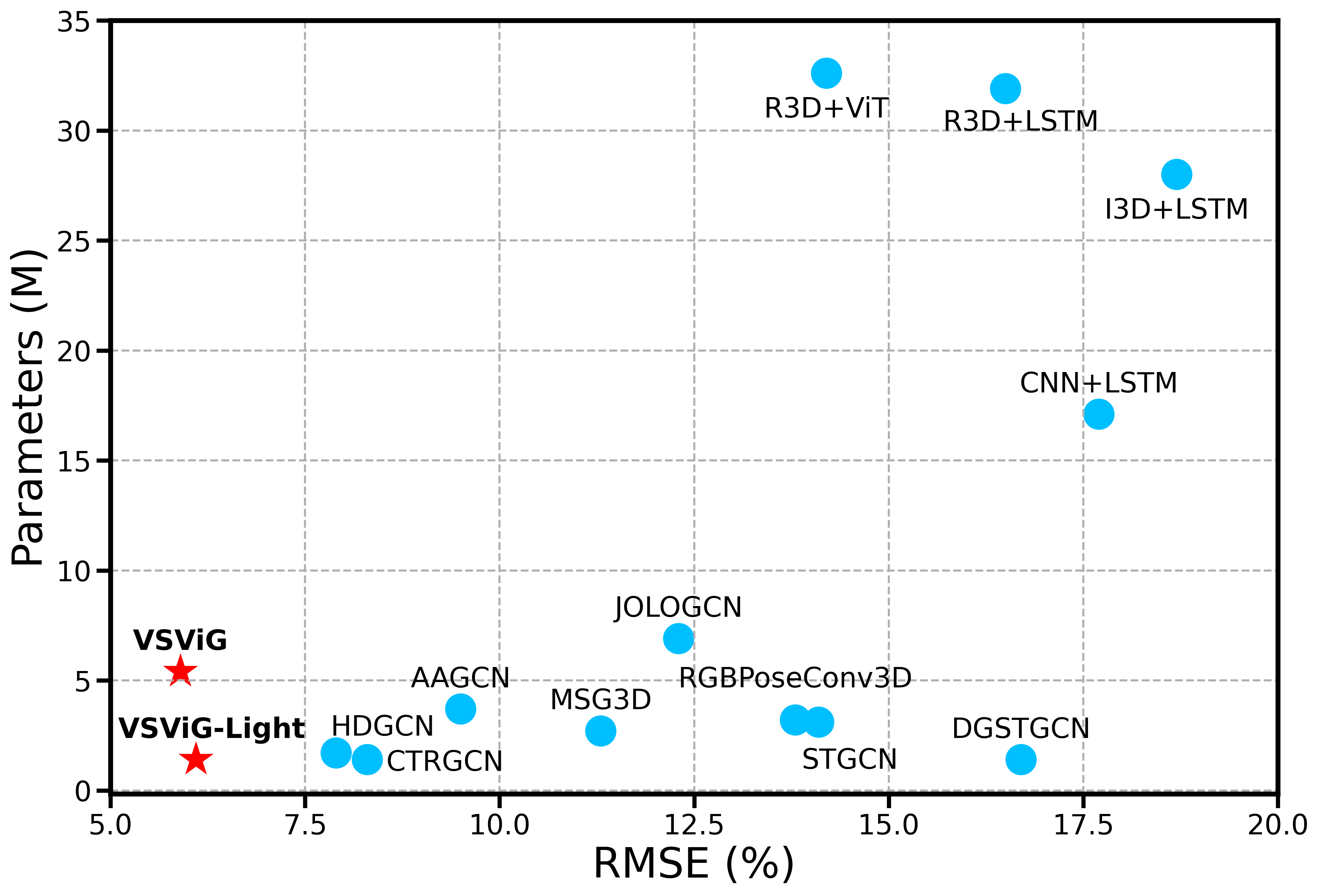}
  \caption{Model comparison of achieved errors and the number of parameters.}
  \label{fig:perform}
\end{figure}

\textbf{(2) Lack of effective analytic tools.} 
Fundamentally, video-based seizure detection is an action recognition task, requiring the analysis of complex seizure-related actions in epileptic patients. 
Prior research indicates that existing methods struggle with achieving short detection latency and are often limited to nocturnal seizures when using raw RGB frames \cite{JBHI-video,van2020automated}. While skeleton-based approaches powered by graph convolutional networks (GCNs) offer several advantages over RGB-based methods \cite{duan2022revisiting,yue2022action,kong2022human}, there also encounter significant limitations in the context of seizure detection.
Firstly, pre-trained pose estimation models \cite{8765346,fang2022alphapose,bazarevsky2020blazepose,liu2021deep,toshev2014deeppose} cannot be applied to epileptic patients' action recognition directly due to the complexity of epileptic patient scenarios, as shown in \cref{fig:visual-1}. Secondly, as shown in \cref{fig:motivate}, there are inherent difficulties in recognizing certain seizure-related actions naively using coordinates as traditional skeleton-based approaches did.

\textbf{(3) Detection latency and false detection rate.}
Detection latency is a critical metric in a seizure detection system, representing the time gap between the real onset of a seizure and the detected onset. A shorter detection latency, known as early seizure detection is highly desirable since it enables timely interventions prior to a serious seizure attack.
Also, FDR is important \cite{tzallas2012automated,pmlr-v56-Thodoroff16}, however, in video-based detection many normal patient behaviors closely resemble seizure-related actions. Previous studies are hard to distinguish them or even directly overlook this metric.

In this study, we propose a novel \textbf{V}ideo-based \textbf{S}eizure detection model via skeleton-based spatiotemporal \textbf{Vi}sion \textbf{G}raph neural network (VSViG) intended for efficient, accurate, and timely real-time scenarios. We address the aforementioned limitations by offering the following contributions: (1) We acquire a dataset of epileptic patient data from hospital EMUs and fine-tune the pose estimation model for epileptic patients; (2) As expressed in the \cref{fig:motivate}, instead of using skeleton-based coordinates, our VSViG model utilizes skeleton-based patch embeddings as inputs to address mentioned challenges. Our proposed model including base and light versions outperforms the state-of-the-art models from the field of action recognition and video-based seizure detection in both accuracy and model size, as shown in \cref{fig:perform}; (3) We define the application as a regression task to generate the probability/likelihood outputs instead of naive binary classification on interictal (healthy) and ictal (unhealthy) status, thereby the model can accurately detect subtle or trending seizure-related actions in low likelihood for early detection propose.

\section{Related Work}
\label{related}

\textbf{Video-based seizure detection.} There have been several efforts paid to video-based seizure detection or jerk detection over the past decade \cite{cuppens2012using,van2020automated,kalitzin2012automatic,geertsema2018automated}, but they neither achieved accurate performance nor supervised patients 24/7 by video monitoring. Recently, several works \cite{JBHI-video,karacsony2022novel,mehta2023privacy,perez2021transfer} made use of advanced deep learning (DL) models to improve the video-based seizure detection performance, however, there are many limitations among these works: (1) \cite{JBHI-video,karacsony2022novel} cannot achieve short detection latency or require more video modality data than RGB; (2) \cite{mehta2023privacy,perez2021transfer} defined the early seizure detection task as a video-based action anticipation task, but they cannot achieve higher accuracy until utilizing at least 1/2 clip of ictal videos which cannot bring short latency, further action anticipation task cannot provide any FDR information.

\textbf{Action recognition.} There are two mainstream strategies for action recognition tasks, one is 3D-CNN for RGB-based action recognition, and the other one is skeleton-based action recognition. 
Many 3D-CNN or ViT architectures \cite{feichtenhofer2019slowfast,feichtenhofer2020x3d,ji20123d,tran2015learning,tran2018closer,chen2022mm,xiang2022spatiotemporal,arnab2021vivit} have been proven to be effective tools for learning spatiotemporal representations of RGB-based video streams, they are widely applied to video understanding and action recognition tasks. Compared to RGB-based approaches with a large number of trainable parameters, skeleton-based approaches perform better in both accuracy and efficiency because they only focus on the skeleton information which is more relevant to the actions, and can alleviate contextual nuisances from raw RGB frames. Several GCN-based approaches \cite{shi2019two,li2019actional,song2020stronger,chen2021channel,yan2018spatial} were proposed to achieve great performance in skeleton-based action recognition. \cite{duan2022revisiting} proposed PoseConv3D to combine 3D-CNN architecture and skeleton-based heatmaps, and proposed its variant RGBPoseConv3D to fuse RGB frames to obtain better performance. However, all aforementioned action recognition models show weaknesses in analyzing seizure-related actions, which include subtle behavioral changes when seizures begin. In this work, we propose VSViG model to address this challenge.

\textbf{Vision graph neural network.} \cite{han2022vision} first proposed vision graph neural networks (ViG) as an efficient and effective alternative backbone for image recognition. Since then many ViG variants \cite{han2023vision,munir2023mobilevig,zhang2023factorized,wu2023pvg} have been proposed to handle various applications. Inspired by ViG, we first propose VSViG to extend ViG to accomplish a skeleton-based action recognition task from seizure-related videos. 

\begin{figure*}[ht]
  \centering
   \includegraphics[width=0.95\textwidth]{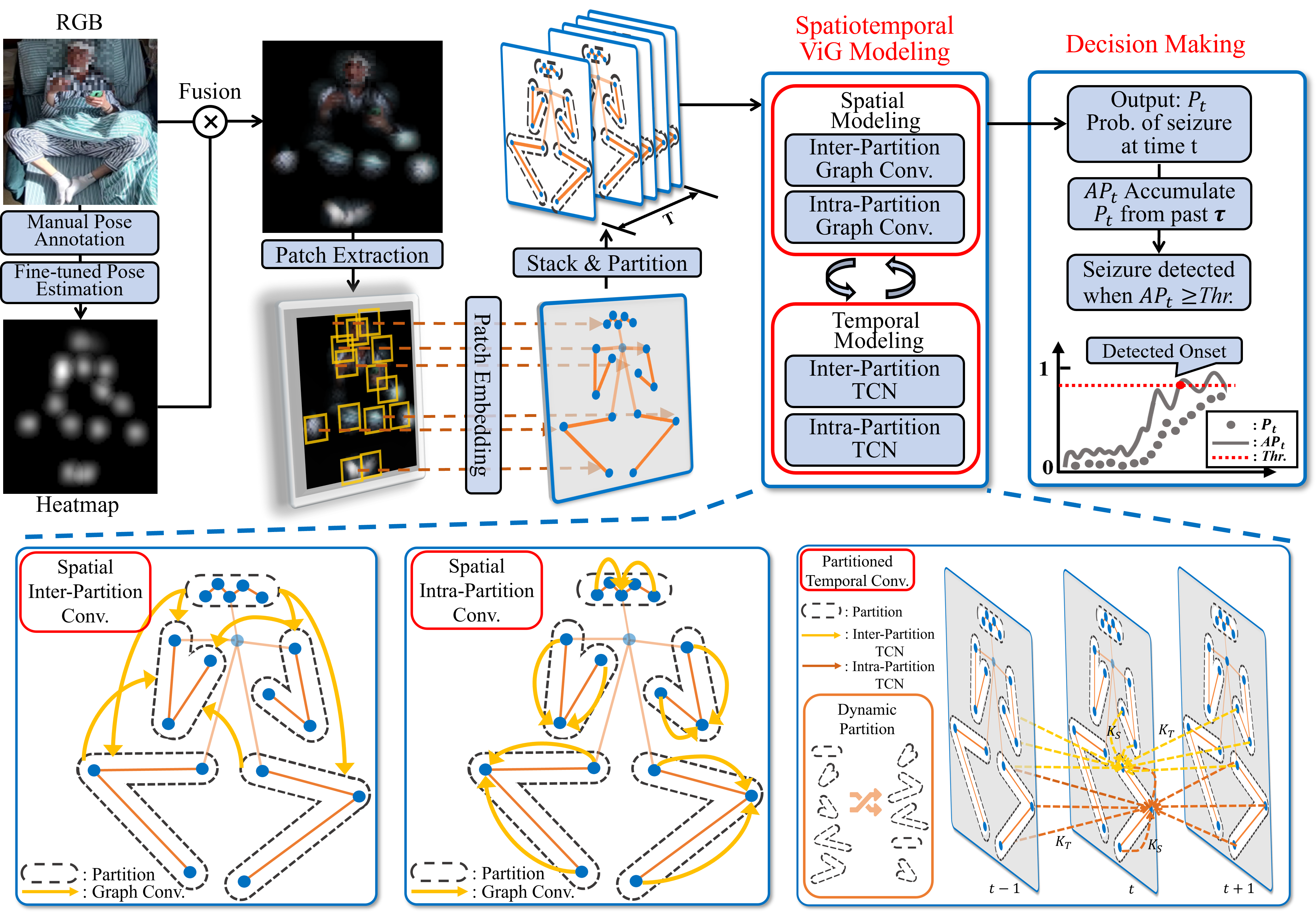}

   \caption{Proposed skeleton-based VSViG framework. Starting from raw RGB frames, we extract skeleton-based patches around each joint by fusing RGB frames and pose heatmaps. Then features from patches are generated by a patch embedding. In spatiotemporal ViG modeling, a partition strategy with proposed inter-, intra-, and dynamic partition operations is used, as shown in the bottom three subfigures.}
   \label{fig:framework}
\end{figure*}

\section{VSViG Framework}
\label{framework}
\cref{fig:framework} presents an overview of skeleton-based VSViG framework. The process begins with a video clip composed of consecutive RGB frames, we fine-tune the pose estimation model to generate joint heatmaps, then fuse these heatmaps and raw frames to construct skeleton-based patches. Before being processed by the VSViG model, these patches are transformed into feature vectors through a trainable patch embedding layer. During the spatiotemporal ViG processing phase, we conduct spatial and temporal modeling respectively, and utilize partition structure to learn inter-partition and intra-partition representations. VSViG model outputs probabilities instead of binary classes for video clips, then a decision-making rule integrating probabilities and accumulative function can achieve shorter detection latency and lower FDR. The subsequent subsections will provide detailed explanations of each step in this process.

\subsection{Pose estimation model fine-tuning}
The VSViG framework relies on a pose estimation algorithm to extract joint heatmaps at first. However, when we tested several mainstream pre-trained pose estimation models on our patient video data, the results were unsatisfactory, which are shown as \cref{fig:visual-1}. These models failed to accurately track the locations of patients and their joints. Therefore, we have to fine-tune the pose estimation model for epileptic patients.

We manually annotated 580 frames containing various behaviors during both ictal periods and healthy status across all patients, then utilized lightweight-openpose \cite{osokin2018lightweight_openpose} as a base model to fine-tune due to its efficiency. Eventually, we achieved a fine-tuned pose estimation model can accurately track patients' joints across all patients, as shown in \cref{fig:visual-1}.

\subsection{Patch extraction and patch embedding}

In custom pose estimation algorithm, each joint is associated with a heatmap generated by a 2D Gaussian map \cite{cao2017realtime}. We make use of this heatmap as a filter to fuse the raw RGB frames, then extract a small patch around each joint from the fused image. 
However, determining the optimal size of the Gaussian maps to capture the regions of interest is a challenge during the training phase of the pose estimation model. To address this, we manually extract the patch $\mathbf{p}_{it}$ of $i$-th ($i=1,2...,N$) joint at frame $t$ by generating a Gaussian map with adjustable $\sigma$ for each joint to filter the raw RGB frame $\textbf{\textit{I}}_{t}$:
\begin{equation}
\mathbf{p}_{it} = \text{exp}(-\frac{(m-x_{it})^2+(n-y_{it})^2}{2\sigma^{2}}) \otimes \textbf{\textit{I}}_{t}(m,n)
\end{equation}
where $|m-x_{i}|\leq H/2$, $|n-y_{i}| \leq W/2$. $(H, W)$, $(m,n)$ and $(x_{it}, y_{it})$ respectively stand for the size of patches, the coordinate of image pixels, and the location of $i$-th joint at frame $t$. The $\sigma$ controls the shape of 2D Gaussian kernels. As a result, we are able to extract a sequence of skeleton-based patches $\mathbf{P} \in \mathbb{R}^{N\times T\times H \times W \times 3}$ from each video clip, then a patch embedding layer transforms patches into 1D feature vectors $\mathbf{X} \in \mathbb{R}^{N\times T\times C}$.

\subsection{Graph construction with partition strategy}
Given a sequence of skeleton-based patch embeddings, we construct an undirected skeleton-based spatiotemporal vision graph as $\mathcal{G} = (\mathcal{V},\mathcal{E})$ on this skeleton sequence across $N$ joints and $T$ stacked frames. Each node in the node set $\mathcal{V} = \left \{ v_{it}|i=1,2,..., N, t=1, 2..., T \right \}$ is associated with feature vectors $\mathbf{x}_{it}\in \mathbb{R}^{C}$.

In this study, we proposed a partitioning strategy to construct graph edges between nodes.
According to Openpose joint template, which contains 18 joints, we focus on 15 joints by excluding three (l\_ear, r\_ear, neck) as redundant. These 15 joints are divided into 5 partitions, each comprising three joints (\textbf{head}: nose, left/right eye; \textbf{right arm}: right wrist/elbow/shoulder; \textbf{right leg}: right hip/knee/ankle; \textbf{left arm}: left wrist/elbow/shoulder; \textbf{left leg}: left hip/knee/ankle). In our VSViG, we conduct both spatial and temporal modeling for the graph. Spatial modeling involves constructing two subsets of edges based on different partition strategies: inter-partition edge set and intra-partition edge set, which are denoted as:
\begin{equation}
\begin{aligned}
    \mathcal{E}_{inter} & =  \left \{ v_{it}v_{jt}|i \in \mathbf{Part}_{p_{1}}, j \in \mathbf{Part}_{p_{2}}, p_{1} \ne p_{2} \right \} \\
    \mathcal{E}_{intra} & = \left \{ v_{it}v_{jt}|(i,j) \in \mathbf{Part}_{p} \right \}
\end{aligned}
\end{equation}
where $p/p_{1}/p_{2} = 1,..., 5$. As shown in the bottom left two schematic figures in \cref{fig:framework}, each node is connected to the nodes from all other partitions in the inter-partition step, and each node is only connected to the other nodes within the same partition in the intra-partition step. The bottom-left two figures of \cref{fig:framework} visualize the graph construction with partitioning strategy.

For temporal modeling, we consider neighbors of each node within a $K_{S} \times K_{T}$ in both temporal and spatial dimensions. The temporal edge set is denoted as $\mathcal{E}_{T} = \left \{ v_{it}v_{j(t+1)} | d(v_{it}, v_{jt}) \leq K_{S}/2, d(v_{it}, v_{i(t+1)}) \leq K_{T}/2 \right \}$, where $d$ is the distance between two nodes. To facilitate the partitioning strategy, we arrange the joints with their partitions in a 1D sequence. Thus, if $K_{S} = 3$, as shown in the bottom-right of \cref{fig:framework}, the middle joint of a partition aggregates only from intra-partition nodes over $K_{T}$ frames, while the border joint of a partition aggregates from inter-partition nodes over $K_{T}$ frames.

\subsection{Partitioning spatiotemporal graph modeling}

\textbf{Spatial Modeling.} The proposed skeleton-based VSViG aims to process the input features $\mathbf{X} \in \mathbb{R}^{N\times T\times C}$ into the probabilities of seizure onset. Starting from input feature vectors, we first conduct spatial modeling to learn the spatial representations between nodes at each frame by inter-partition and intra-partition graph convolution operation as $\mathcal{G}^{'} = F_{intra}(F_{inter}(\mathcal{G},\mathcal{W}_{inter}), \mathcal{W}_{intra})$, where $F$ is graph convolution operation, specifically we adopt max-relative graph convolution \cite{li2019deepgcns,han2022vision} to aggregate and update the nodes with fully-connected (FC) layer and activation function:
\begin{equation}
\label{eq:mr}
    \mathbf{x}^{'}_{it} = \sigma(Linear([\mathbf{x}_{it}, max({\mathbf{x}_{jt}-\mathbf{x}_{it}}| j \in \mathcal{N}^{t}(\mathbf{x}_{it}))]))
\end{equation}
where $\mathcal{N}^{t}(\mathbf{x}_{it})$ is neighbor nodes of $\mathbf{x}_{it}$ at frame $t$, $Linear$ is a  FC layer with learnable weights and $\sigma$ is activation function, \eg ReLU. The only difference between $F_{inter}$ and $F_{intra}$ depends on $\mathcal{N}^{t}(\mathbf{x}_{it})$ is from $\mathcal{E}_{inter}$ or $\mathcal{E}_{intra}$. Given input feature vectors at frame $t$, denoted as $\mathbf{X}_{t} \in \mathbb{R}^{N\times C}$, the graph convolution processing as \cref{eq:mr} can be denoted as $\mathbf{X}^{'}_{t} =  MRGC(\mathbf{X}_{t})$, then the partitioning spatial graph processing module in VSViG would be:
\begin{equation}
\begin{aligned}
    \mathbf{X}_{t}^{''} \ & = \sigma(MRGC(\mathbf{X}_{t}W_{in\_inter})W_{out\_inter} + \mathbf{X}) \\
    \mathbf{X}_{t}^{'} \ & =  \sigma(MRGC(\mathbf{X}_{t}^{''}W_{in\_intra})W_{out\_intra} + \mathbf{X})
\end{aligned}
\end{equation}
We add original input $\mathbf{X}_{t}$ as a residual component to both two partitioning spatial modeling, which is intended to avoid smoothing and keep diversity of features during the model training \cite{he2016deep,dang2021msr}. We keep output dimension of every learnable layer is same as input dimension, so that we achieve output feature $\mathbf{X}^{'} \in \mathbb{R}^{N\times T\times C}$ when we stack $T$ frames $\mathbf{X}^{'}_{t} \in \mathbb{R}^{N\times C}$ after spatial modeling.

\textbf{Temporal Modeling.} Given the dimension of output features after spatial modeling is $N\times T\times C$, and we order the partitions (containing joints) in a 1D sequence, so that we can naturally consider the input for the temporal modeling as a 3D volume, denoted as $\mathbf{X} \in \mathbb{R}^{1\times N\times T\times C}$. According to the construction of graph in temporal modeling $\mathcal{G} = (\mathcal{V}, \mathcal{E}_{T})$, we can adopt convolution operation to aggregate and update nodes:
\begin{equation}
\label{eq:conv3d}
    \mathbf{x}^{'}_{i} = \sum_{j}^{1}\sum_{j}^{K_{S}}\sum_{j}^{K_{T}}w_{ij}\mathbf{x}_{i}
\end{equation}
 Inspired by \cite{duan2022revisiting,yan2018spatial,he2016deep}, we utilize residual 3D-CNN architecture, denoted as $Conv$, to conduct temporal graph modeling as \cref{eq:conv3d} for its simplicity:
\begin{equation}
    \mathbf{X}^{'} = Conv{(1)}(Conv{(1,K_{S},K_{T})}(Conv{(1)}(\mathbf{X}))) + \mathbf{X}
\end{equation}
where $(1)$ and $(1,K_{S},K_{T})$ are kernel size for the 3D convolution operation.

\begin{figure}[t]
  \centering
   \includegraphics[width=0.6\columnwidth]{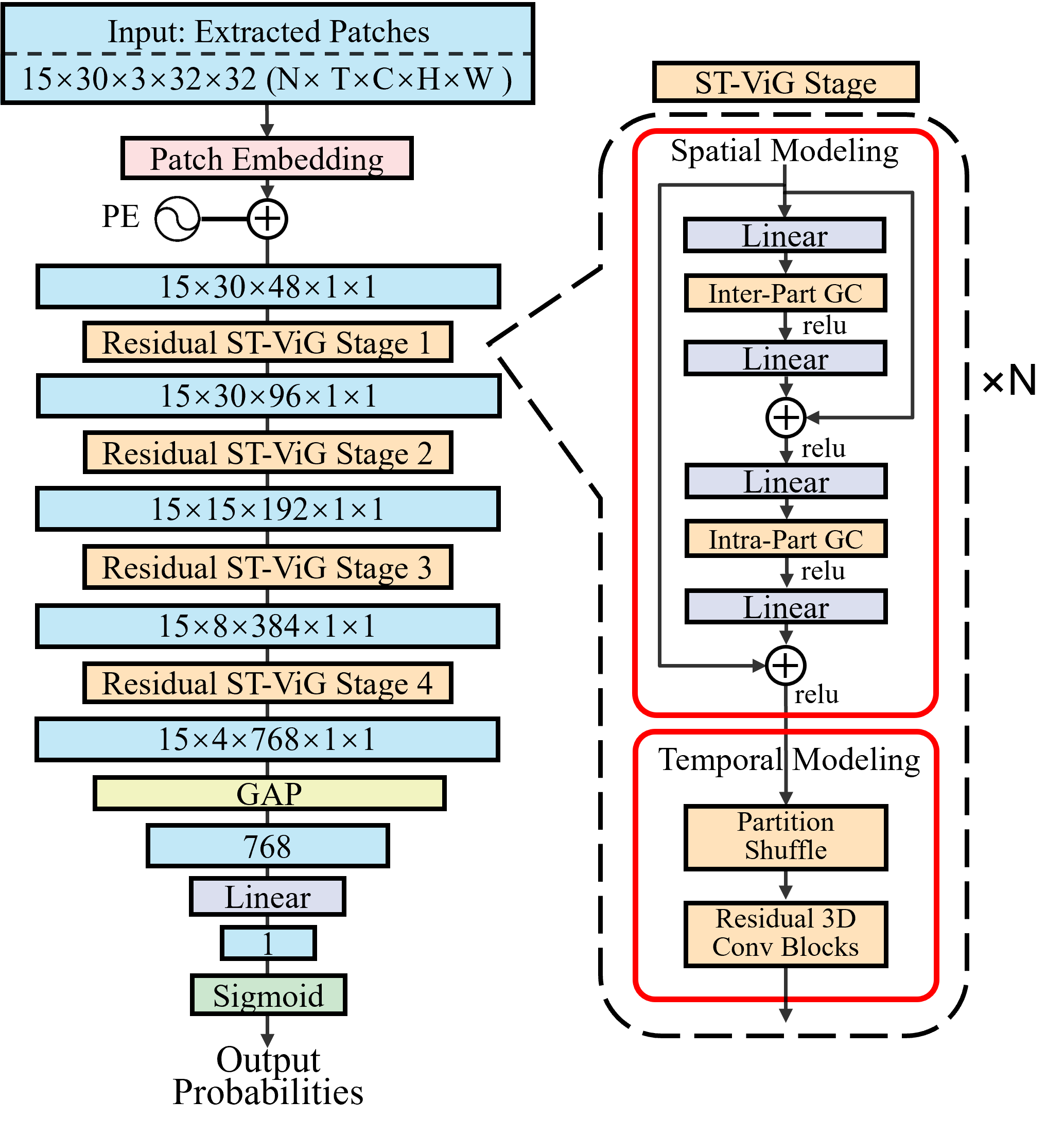}

   \caption{VSViG architecture. VSViG consists of four residual spatiotemporal (ST) ViG stages and each stage contains several proposed spatial and temporal modeling layers. $N,T,C$ denotes the number of joints, frames and channels, and $H,W$ stand for the height and width of extracted patches.}
   \label{fig:layers}
\end{figure}

\subsection{VSViG network architecture}

\cref{fig:layers} shows the detailed structure of proposed VSViG network including the shape of features at each layer. We keep feature size in spatial modeling, and implement downsampling at the last layer of each residual temporal block. After four residual spatiotemporal ViG stages, we make use of global average pooling to generate a 1D feature vector, then one FC layer and sigmoid function are connected to generate output probabilities. 
In spatial modeling, $E=12$ and $E=2$ represent the number of edges (neighbors) in inter-partition and intra-partition graph convolution, respectively. The spatial modeling phase always keeps the output channels unchanged. In temporal modeling, we set $K_{S} = 3$ for simultaneous implementation of inter-partition and intra-partition temporal convolution network (TCN) operation, and set $K_{T}=3$ for considering only one frame before and after the current frame. The residual 3D-CNN block is used to expand feature dimension and downsample temporal frames. 

We propose two versions of VSViG models: VSViG and VSViG-Light respectively based on different model depths and feature expansions, they are designed for accurate or efficient purposes. Their details are illustrated in \textbf{Appendix}.

\textbf{Positional embedding.} In previous studies, either skeleton-based STGCNs which input coordinate triplets or RGB-/Heatmap-based methods made use of specific joint coordinates or whole body context with global positional information. In this work, however, input patches in 1D sequence do not provide any positional information. Thus, we add a positional embedding to each node feature vector. Here we propose 3 different ways to inject positional information: (1) Add learnable weight: $\mathbf{x}_{it} = \mathbf{x}_{it} + \mathbf{e}_{it}$; (2) Concatenation with joint coordinates: $\mathbf{x}_{it} = [\mathbf{x}_{it}, (x_{it},y_{it})]$; (3) Add embeddings of joint coordinates: $\mathbf{x}_{it} = \mathbf{x}_{it} + Stem(x_{it},y_{it})$;

\textbf{Dynamic partitions.} Since the partitions are arranged as a 1D sequence, each partition is only connected to 1 or 2 adjacent partitions, it cannot consider all other partitions as spatial modeling does. Thus, we obtain the dynamic partitions by conducting partition shuffle before temporal graph updating. It is noted that we only shuffle the partitions, and the order of joints within a partition is unchanged. 

\subsection{Seizure onset decision-making}
\label{subsec:decision}
Inspired by \cite{XU2024121359}, after generating output probabilities $P_{t}$ by proposed skeleton-based VSViG model for video clips at time t, we can accumulate a period $\tau$ of previously detected probabilities as accumulative probabilities $AP_{t} = \sum_{i=t-\tau}^{t} P_{i}$ with detection rate $r$, then make a seizure onset decision at time $t_{onset}$ when accumulative probabilities reach a decision threshold $DT$: $t_{onset} = AP_{t} > DT$. We measure the distance between $t_{onset}$ and EEG onset as latency of EEG onset $L_{EO}$, and the distance between $t_{onset}$ and clinical onset as latency of clinical onset $L_{CO}$.

\begin{figure}[t]

  \centering
   \includegraphics[width=0.5\columnwidth]{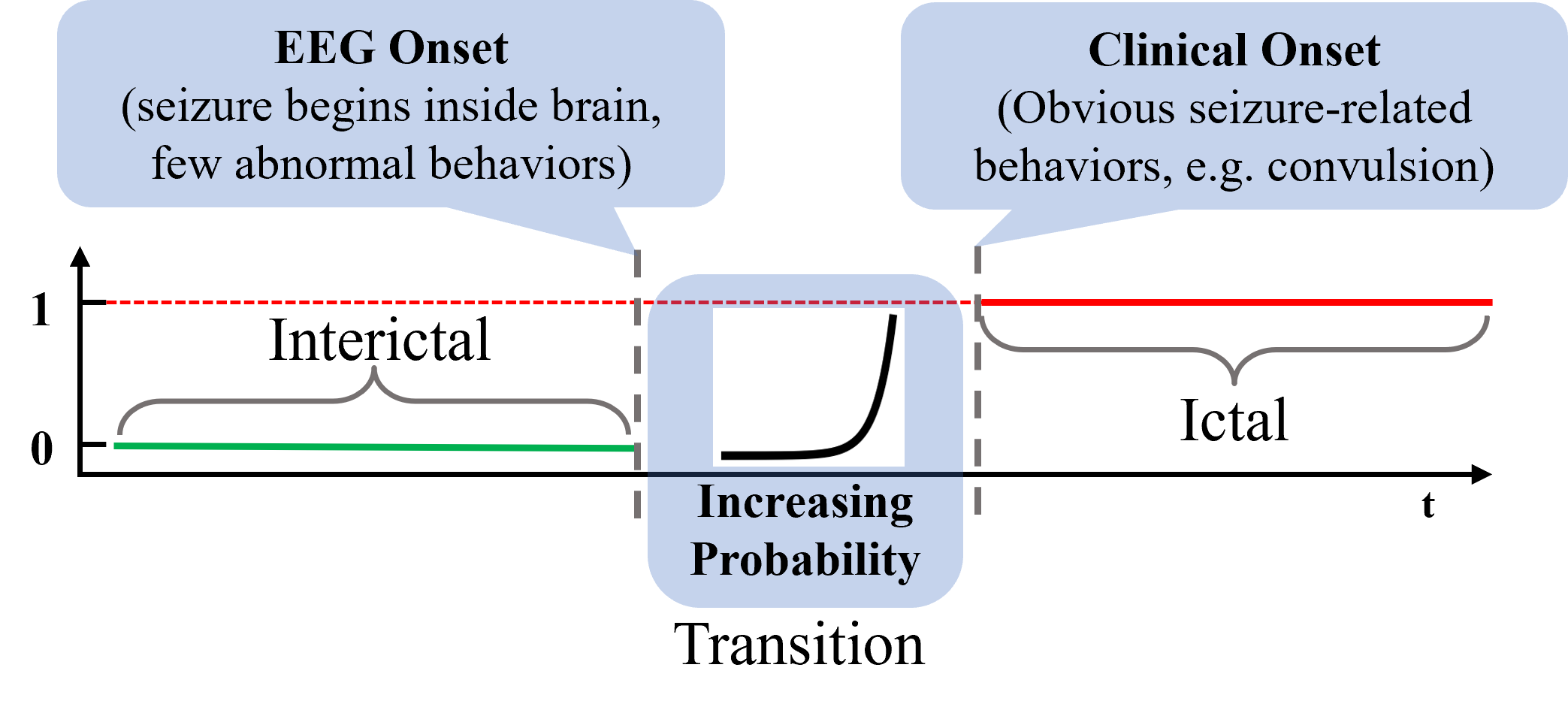}

   \caption{Data labeling for a regression-based task. For each seizure, a video recording is categorized into 3 different periods: interictal (label: 0), ictal (label: 1), and transition (label: 0 to 1 in exponential function according to clinical phenomenon).}
   \label{fig:onset}
\end{figure}

\section{Experiments}
\label{experiments}

\subsection{Dataset}
We acquire surveillance video data of epileptic patients from the local hospital, this dataset includes 14 epileptic patients with 33 focal or generalized tonic-clonic seizures. As for each seizure, we utilize all ictal periods, and 5-30 min interictal periods before the EEG onset. We segment the raw video recording into video clips with 5 s for model training and manually label the video clips as likelihood for regression task according to \cref{fig:onset}. Naive binary classification of healthy and unhealthy status cannot accurately recognize the early occurrence of seizures, the introduced increasing probability way is consistent with the likelihood of real seizure attack risk. And an exponential way is more similar to the clinical phenomenon that the closer time to the clinical onset, the higher risk is obtained. We also test other functions to build increasing probability labels for this regression task, which is shown in \textbf{Appendix}.

Eventually, 2.78 h interictal period, 0.16 h transition period and 0.35 h ictal period video are obtained for the model training, and overlapping extraction operation is utilized to overcome the issue of data imbalance. Then we split 70\% of each period for training, and the rest 30\% for validation. More details information about the dataset is illustrated in \textbf{Appendix}.

\textbf{Performance metric.} In this paper, we define a regression task intended for early seizure detection, RMSE metric measured between true video clip and predictive video clip is used to compare the model performance.

\textbf{Data usage.} The data collection from hospital is carefully conducted and the use of data was approved by hospital Institutional Review Board. We will release this dataset when all sensitive and private information is removed. Code and fine-tuned pose estimation model will be also released.

\subsection{Experimental setting}
\label{subsec:setting}
We manually set a size of $32\times 32 \ (H \times W)$ for extracted patches based on the 1920$\times$ 1080 raw frame resolution, and $\sigma$ of gaussian kernel is $0.3$. All generated outputs are connected with a Sigmoid function to map the outputs as probabilities from 0 to 1, the activation function used anywhere else is chosen as ReLU, and mean square error loss function is used to train the model. During the training phase, epoch and batch size are respectively set to $200$ and $32$, we choose Adam optimizer with 1e-4 learning rate with 0.1 gamma decay every 40 epochs and 1e-6 weight decay.

\subsection{Seizure-related action recognition performance}
According to \Cref{tab:compare}, VSViG and VSViG-Light respectively obtain $5.9\%$ and $6.1\%$ errors across all patients, which outperform all previous state-of-the-art approaches. Furthermore, VSViG-Light only obtains 0.44G FLOPs and 1.4M model size which is desirable for real-time deployment. 

We also reproduce several state-of-the-art action recognition and video-based seizure detection models to make comparisons with proposed VSViG model. There are four selective state-of-the-art video-based seizure detection approaches, they are all RGB-based strategies utilizing a CNN-based model to extract features that are followed by a sequence-based model. In terms of skeleton-based action recognition models, several state-of-the-art GCN-based models are chosen, and one RGB+skeleton fusion method is also considered. 

\begin{table*}[t]
  \caption{Model comparisons. We reproduce state-of-the-art approaches to make comparisons with proposed VSViG. (OF: optical flow, $\clubsuit$: Video-based seizure detection models, $\spadesuit$: Skeleton-based action recognition models, $\bigstar$: This work)}
\centering
\footnotesize
  \begin{tabular}{l|c c c c c}
    \specialrule{.1em}{.05em}{.05em} 
    \makecell[c]{Method} & Input & RMSE(\%)& GFLOPs  & Param.(M)\\
    \specialrule{.1em}{.05em}{.05em}

    $\clubsuit$ I3D+LSTM\cite{karacsony2022novel} & RGB& 18.7& 57.3& 28.0\\
    $\clubsuit$ CNN+LSTM\cite{JBHI-video} & RGB& 17.7& 5.0& 17.1\\
    $\clubsuit$ R3D+LSTM\cite{perez2021transfer}& RGB& 16.5& 163.1 & 31.9\\
    $\clubsuit$ R3D+ViT\cite{mehta2023privacy}& RGB& 14.2& 163.1& 32.6\\
    \specialrule{.1em}{.05em}{.05em} 
    
    $\spadesuit$ DGSTGCN \cite{duan2022dg} & Skeleton& 16.7  &0.50 & 1.4\\
    $\spadesuit$ STGCN \cite{yan2018spatial} & Skeleton&14.1  & 1.22 & 3.1\\
    $\spadesuit$ RGBPoseConv3D \cite{duan2022revisiting} & RGB+Skeleton & 13.8&12.63 & 3.2\\
    $\spadesuit$ JOLOGCN \cite{cai2021jolo} & OF in Patches & 12.3&37.30 & 6.9\\
    
    $\spadesuit$ MSG3D \cite{liu2020disentangling} & Skeleton&11.3  &1.82& 2.7\\
    $\spadesuit$ AAGCN \cite{shi2020skeleton} & Skeleton& 9.5  &1.38& 3.7\\
    $\spadesuit$ CTRGCN \cite{chen2021channel} & Skeleton& 8.3 & 0.62& 1.4\\
    $\spadesuit$ HDGCN \cite{Lee_2023_ICCV} & Skeleton& 7.9 & 1.60& 1.7\\
    \specialrule{.1em}{.05em}{.05em} 
    $\bigstar$ VSViG-Light & Patches & 6.1 & \textbf{0.44}& \textbf{1.4}\\
    $\bigstar$ VSViG &  Patches & \textbf{5.9}& 1.76& 5.4M\\
    \specialrule{.1em}{.05em}{.05em} 
  \end{tabular}

  \label{tab:compare}
\end{table*}

We can see that RGB-based approaches perform worse in both error and FLOPs than skeleton-based approaches, it is consistent with RGB-based approach's drawbacks: large model size and can easily be disturbed by nuisances. Furthermore, several previous video-based seizure detection model rely on the pre-trained RGB-based 3D CNN model to extract features, which extremely increases the computational resources.  
As for skeleton-based approaches, most perform around $10\%$ error with lower FLOPs, and CTRGCN performs the best among all skeleton-based approaches.
These results underscore the superiority of our skeleton-based patch embedding VSViG models over both RGB-based and skeleton-based GCN approaches

\subsection{Visualization}
\begin{figure}[t]
  \centering
   \includegraphics[width=\textwidth]{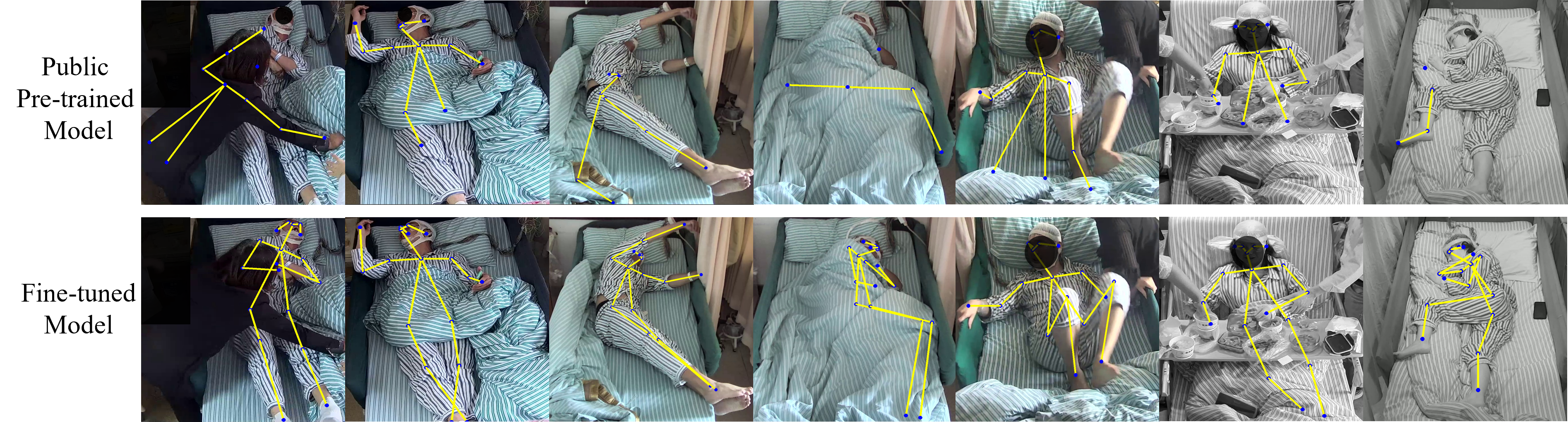}

   \caption{Pose estimation performance comparison between public pre-trained model and fine-tuned model. Black circles are used to mask sensitive information.}
   \label{fig:visual-1}
\end{figure}

\begin{figure}[t]
  \centering
   \includegraphics[width=\textwidth]{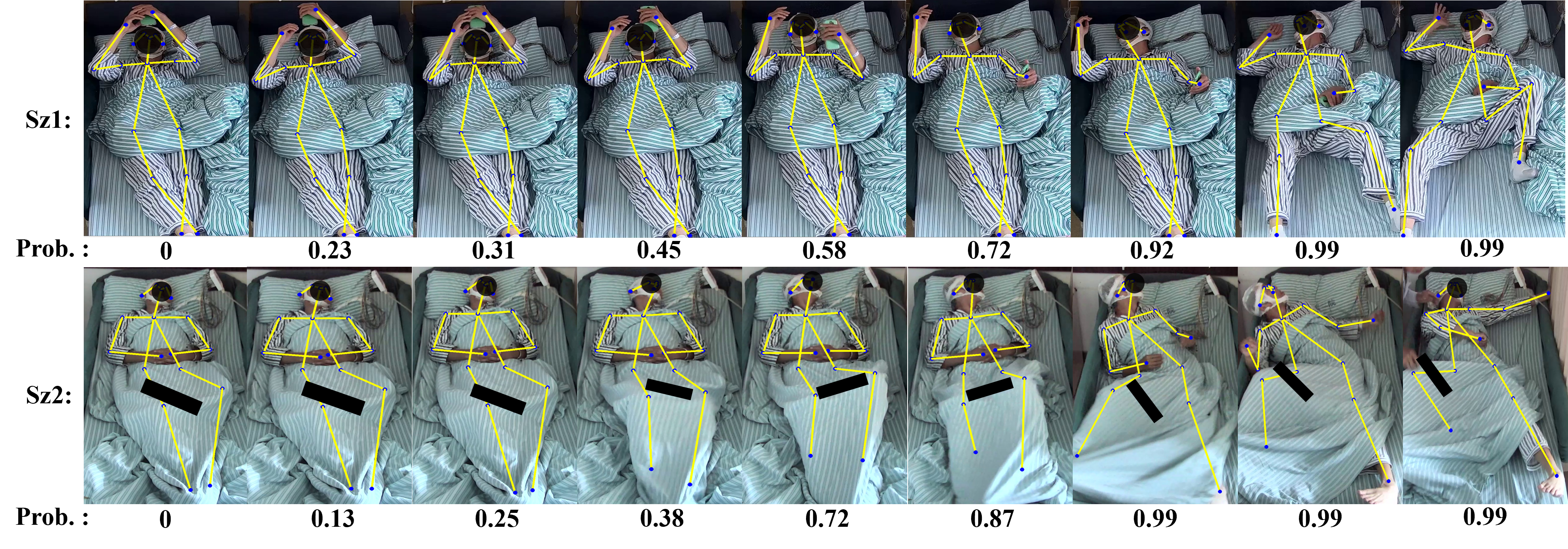}

   \caption{Predictive probabilities/likelihoods of video clips from two different seizures are obtained by VSViG model. Black boxes are used to keep privacy.}
   \label{fig:visual-2}
\end{figure}

We visualize the pose estimation performance comparison between our fine-tuned model and the public pre-trained model, as shown in \cref{fig:visual-1}, thereby demonstrating the limitation of public pre-trained model for tracking patient skeletons.
Secondly, we present the visualization of predictive probability of each video clip obtained by VSViG model in a successive seizure from two different seizures, which is shown as \cref{fig:visual-2}.

\subsection{Ablation study}
\label{subsec:ablation}
We conduct ablation studies with VSViG to evaluate the effects of several modules and settings on the model performance, all shown values in \Cref{tab:ablation} are RMSE errors on the validation set.

\textbf{Effect of spatial modeling module.} There are several spatial modeling modules in proposed VSViG, we conduct an ablation study here to prove the effectiveness of inter-partition and intra-partition modules. We can see from \Cref{tab:ablation-spatial} that proposed inter-intra module outperforms other conditions. Inter-partition and intra-partition fundamentally extract coarse and fine features respectively, and Inter-Intra strategy is consistent with coarse-to-fine feature extraction that can learn the representations better.

\textbf{Effect of temporal modeling kernel size.} As for the temporal modeling, a kernel size of $(3, 3)$ stands for $K_{S}$, $K_{T}$ in TCN operation, $K_{S}=3$ means spatial neighbor joints are also considered, meanwhile $K_{S}=1$ is not. And we also compare 3 and 5 convolution size in temporal dimension $K_{T}$. The result in \Cref{tab:ablation-temporal} shows that considering neighbor nodes and $K_{T}=3$ performs best.  

\textbf{Effect of dynamic partition.} In \Cref{tab:ablation-pos}, we can see that Stem positional embedding way with dynamic partition achieves the best performance. Dynamic partition strategy shows better performance, which indicates it can effectively enhance the model to learn relationships between different partitions, thereby understanding the seizure-related actions. 

\textbf{Effect of positional embedding.} Also, positional information can bring better performance for VSViG model, and Stem way outperforms all conditions. Further, no positional embedding with dynamic partition can also achieve a satisfactory result compared with other GCN-based approaches that utilize coordinates as input, this indicates that proposed VSViG model has the capability to complete an action recognition task without any specific or global positional information. In our point of view, the reason is that inter-partition and intra-partition representations contain enough information for action recognition no matter how these partitions are ordered or where they are.

\textbf{Effect of gaussian kernel size.} During the phase of fine-tuning lightweight-openpose, $\sigma$ size should be set in advance. Smaller size brings less RGB information in extracted patches, meanwhile larger size brings more contextual nuisances. In this ablation study \Cref{tab:ablation-sigma}, 0.3 relative to the patch size is the best. It is noted that the chosen patch size depends on the video resolution and size of patient in the video, $32\times32$ is suitable for this scenario.

\begin{table}[t]
    \scriptsize
    \caption{Ablation studies on VSViG. All experiments are conducted with VSViG version, and values are RMSE.} 
    \label{tab:ablation}
    \vspace{-0.6cm}
    \begin{subtable}{.5\textwidth}
      \centering
        \caption{Spatial modeling module.}
        \vspace{-0.5cm}
        \begin{tabular}[t]{cc|c}
    \specialrule{.1em}{.05em}{.05em} 
    \quad Inter \quad & \quad Intra \quad & \quad RMSE \quad \\
    \specialrule{.1em}{.05em}{.05em}
    \quad \ding{55} \quad &\quad \ding{55} \quad&\quad 13.0 \quad \\
    \quad \ding{51} \quad &  \quad \ding{55} \quad &\quad 11.7 \quad \\
   \quad \ding{55} \quad&\quad \ding{51} \quad&\quad 9.2 \quad \\
    \quad \ding{51} \quad &\quad \ding{51} \quad&\quad \textbf{5.9} \quad \\
    \specialrule{.1em}{.05em}{.05em}
\end{tabular}

\label{tab:ablation-spatial}
    \end{subtable}%
    \begin{subtable}{.5\textwidth}
      \centering
        \caption{Temporal modeling kernel size.}
        \vspace{-0.5cm}
        \begin{tabular}[t]{c|c}
    \specialrule{.1em}{.05em}{.05em} 
    \quad $(K_{S},K_{T})$ \quad & \quad RMSE \quad \\
    \specialrule{.1em}{.05em}{.05em}
    \quad (1, 3) \quad &\quad 9.9 \quad \\
    \quad (1, 5) \quad &\quad 10.8 \quad \\
    \quad (3, 3) \quad &\quad \textbf{5.9} \quad \\
    \quad (3, 5) \quad &\quad 6.5 \quad \\
    \specialrule{.1em}{.05em}{.05em}
\end{tabular}

\label{tab:ablation-temporal}

    \end{subtable}

    \vspace{0.01cm}

        \begin{subtable}{.5\textwidth}
      \centering
        \caption{Positional embedding and dynamic partition.}
        \vspace{-0.15cm}
                 \begin{tabular}{c|cc}
    \specialrule{.1em}{.05em}{.05em} 
    \multirow{2}{5em}{\centering Positional embedding}   & \multicolumn{2}{c}{Dynamic partition} \\
     & \multicolumn{2}{c}{w/o \qquad w.} \\
    \specialrule{.1em}{.05em}{.05em} 
    None  & \multicolumn{2}{c}{16.9 \quad \ 9.1} \\

    Cat. & \multicolumn{2}{c}{14.8 \quad \ 8.9} \\

    Learnable & \multicolumn{2}{c}{13.4 \quad \ 8.6} \\

    Stem (x, y) & \multicolumn{2}{c}{\ 8.4 \quad \ \ \textbf{5.9}} \\
    \specialrule{.1em}{.05em}{.05em} 
  \end{tabular}

\label{tab:ablation-pos}
    \end{subtable}%
    \begin{subtable}{.5\textwidth}
      \centering
        \caption{Gaussian kernel ($\sigma$) size.}
        \vspace{-0.15cm}
  \begin{tabular}{c|c}
    \specialrule{.1em}{.05em}{.05em} 
    \quad $\sigma$ \quad & \quad RMSE \quad \\
    \specialrule{.1em}{.05em}{.05em} 
    \quad 0.1 \quad&  \quad 10.1 \quad \\
    \quad 0.2 \quad&  \quad 8.5 \quad \\
    \quad 0.3 \quad&  \quad \textbf{5.9} \quad \\
    \quad 0.4 \quad&  \quad 8.6 \quad \\
    \quad 0.5 \quad&  \quad 7.6 \quad \\
    
    \specialrule{.1em}{.05em}{.05em}
  \end{tabular}

\label{tab:ablation-sigma}
    \end{subtable} 
\end{table}

\begin{figure}[!t]
  \centering
  \begin{subfigure}{0.24\columnwidth}
    \centering
    \includegraphics[width=\columnwidth]{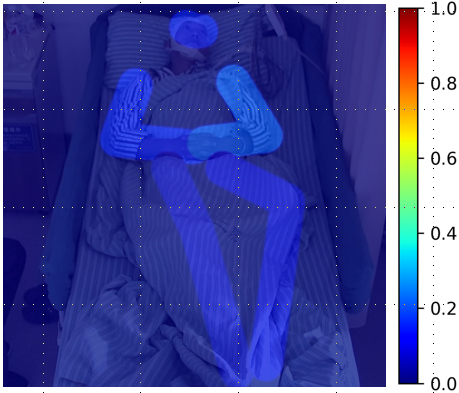}
    \caption{}
    \label{fig:interp-a}
  \end{subfigure}
  \hfill
  \begin{subfigure}{0.24\columnwidth}
    \centering
    \includegraphics[width=\columnwidth]{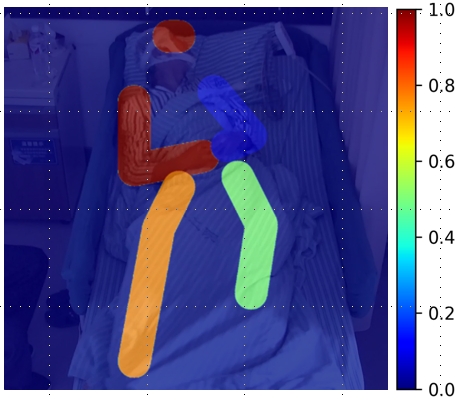}
    \caption{}
    \label{fig:interp-b}
  \end{subfigure}
    \hfill
  \begin{subfigure}{0.24\columnwidth}
    \centering
    \includegraphics[width=\columnwidth]{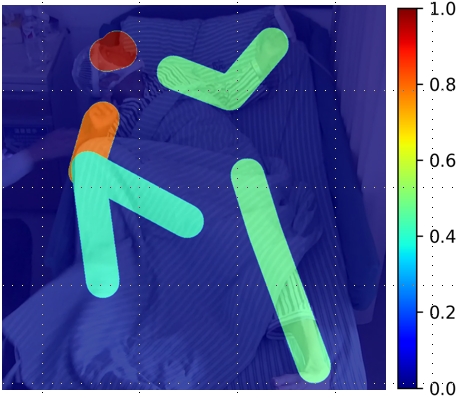}
    \caption{}
    \label{fig:interp-c}
  \end{subfigure}
    \hfill
  \begin{subfigure}{0.24\columnwidth}
    \centering
    \includegraphics[width=\columnwidth]{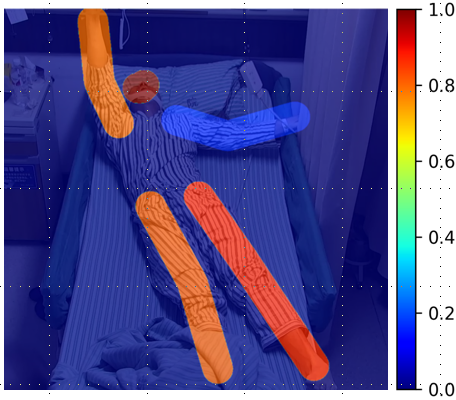}
    \caption{}
    \label{fig:interp-d}
  \end{subfigure}
  \caption{Occlusion maps for model interpretability. Each image represents the end frame of the video clip. Redder partitions mean more salient to the seizure-related actions. (a)-(d) present a patient suffering a seizure from healthy status to convulsion.}
  \label{fig:interp}
\end{figure}

\subsection{Model interpretability}
We provide interpretability with occlusion maps \cite{zeiler2013visualizing} to demonstrate the capabilities of our VSViG model to learn representations of seizure-related actions within different partitions. We successively take 4 video clips of a patient from healthy status to convulsion, as shown in \cref{fig:interp}, each image represents the end frame of corresponding clips, higher value (redder) means the partition is more salient to seizure-related action, and lower value (bluer) means less relevant.
We can see that, initially, when the patient is lying on the bed, all partitions are shown as less salient in \cref{fig:interp-a}. As the seizure begins, the patient starts to exhibit abnormal movements, such as turning the head and moving the right arm.  Correspondingly, these partitions appear redder in \cref{fig:interp-b} and \cref{fig:interp-c}, indicating their increased relevance to the seizure activity. Several seconds later, as the patient begins jerking, most partitions are shown as more salient in \cref{fig:interp-d}. 

\subsection{Make seizure onset decision}
After obtaining probabilities from consecutive video clips by VSViG model, we set up $\tau=3\ s, r=0.5\ s$ as mentioned in \Cref{subsec:decision} to calculate $L_{EO}, L_{CO}$ and FDR across all seizures. We also evaluate the effect of accumulative strategy on the detected latency, as shown in \Cref{tab:latency}. We can see that performance without accumulative strategy is satisfactory based on the accurate VSViG model, but accumulative strategy with $DT = 0.3$ can obtain much better performance where $5.1\ s \ L_{EO}$, -$13.1\ s \ L_{CO}$, which means seizure can be detected only $5.1\ s$ after seizure begins inside brain, and can alarm the seizure coming $13.1\ s$ before patients tend to exhibit serious convulsions. And FDR of 0 is also obtained. The specific results of every patient are shown in \textbf{Appendix}.

\begin{table}[ht]
  \caption{Seizure detection performance. We evaluate the effect of different DTs and the proposed accumulative strategy for detected latency. $DT$: decision threshold, $L_{EO}$: latency of EEG onset $L_{CO}$: latency of clinical onset, FDR: false detection rate.}
\centering
\footnotesize
  \begin{tabular}{ccccc}
  \specialrule{.1em}{.05em}{.05em} 
   \multirow{2}{2em}{\centering $DT$} & \multirow{2}{2.5em}{\centering FDR} & \multirow{2}{4em}{\centering Latency} & \multicolumn{2}{c}{Accumulative} \\
   & & &w/o & w. \\
  \specialrule{.1em}{.05em}{.05em} 
  \multirow{2}{2em}{\centering 0.3} & \multirow{2}{2.5em}{\centering 0} & $L_{EO}$ & 11.4s & \textbf{5.1s} \\
  & & $L_{CO}$ & -6.8s & \textbf{-13.1s} \\
  \specialrule{.1em}{.05em}{.05em} 
  \end{tabular}

  \label{tab:latency}
  
\end{table}

\section{Conclusion}
\label{conclusion}

In this work, 
we proposed a VSViG model utilizing skeleton-based patch embeddings to recognize the epileptic patients' actions effectively, experiments indicate proposed VSViG model outperforms state-of-the-art action recognition approaches.
VSViG provides an accurate, efficient, and timely solution for video-based early seizure detection to fundamentally help epileptic patients.

In the future, we will collect more patient video data with various types of seizures to train a more generalized model for a large number of epileptic patients. And action recognition-based VSViG model can also be extended to detect other movement-based diseases, \eg Parkinson's disease, fall detection.

\section*{Acknowledgement}
This work was supported by STI2030-Major Projects
(2022ZD0208805), ”Pioneer” and ”Leading Goose” R\&D Program of
Zhejiang (2024C03002), and the Key Project of Westlake Institute for
Optoelectronics (Grant No. 2023GD004).


%
%
\bibliographystyle{splncs04}
\bibliography{main}

\clearpage
\setcounter{page}{1}
\section*{A. Data details}
\textbf{Data acquisition.} We acquire surveillance video data of epileptic patients from a hospital, this dataset includes 14 epileptic patients with 33 tonic-clonic seizures. Patients in the hospital EMUs are supervised by Bosch NDP-4502-Z12 camera with 1920 $\times$ 1080 resolution 24/7. We extract successive frames as video clips for real-time processing. Each video clip spans a duration of $5s$ with the original rate of 30 FPS. To enhance the efficiency of our real-time analysis, we apply fixed stride sampling to reduce the frame count from the original 150 frames to 30 frames.

\textbf{Data labeling.} How to label the video clips is a challenge in video-based early seizure detection. Typically, EEG-based seizure detection is treated as a binary classification task, distinguishing between interictal (pre-seizure) and ictal (during-seizure) periods. Medical experts annotate the moment from interictal to ictal periods in EEG recordings as the EEG onset. 
However, in video analysis, the transition from normal to seizure behavior is not as immediate. For several seconds (ranging from 1 to 30 seconds) after the EEG onset, a patient's behavior may appear only slightly abnormal or nearly normal. We refer to this phase as the transition period. The moment when patients start exhibiting clearly abnormal actions, such as jerking, convulsion, or stiffening, is defined as clinical onset.
It is straightforward to label video clips as healthy status (0) before the EEG onset and as seizure status (1) after the clinical onset. These labels (either 0 or 1) can be interpreted as the probability or risk of seizure onset.
In terms of the transition period, it is hard to assign a precise probability to each clip. 
Clinically, it's observed in the transition period that patients are more likely to appear healthy closer to the EEG onset and more likely to exhibit abnormal behaviors as the time approaches the clinical onset. 
To address this, we use an exponential function to assign increasing probabilities to video clips during the transition period, ranging from 0 to 1. Other options, sigmoid and linear functions, are compared to exponential way in \Cref{tab:function}. It is noted that the probabilities of video clips depend on the end frame of video clips lying in which period. Consequently, we define this task as a probability regression task rather than a traditional classification task.

\begin{table}[ht]
\centering
\footnotesize
  \caption{Comparisons of different increasing probability functions used for labeling transition period.}
\setlength{\tabcolsep}{5mm}
  \begin{tabular}{c c c c}
  \specialrule{.1em}{.05em}{.05em} 
   &Linear & Sigmoid & Exponential\\
  \specialrule{.1em}{.05em}{.05em}
  RMSE &6.7\%&9.4\%&\textbf{5.9\%}\\
  \specialrule{.1em}{.05em}{.05em} 
  \end{tabular}
  \label{tab:function}
  
\end{table}

\textbf{Data splitting.} For each seizure event, we categorize the video segments based on their timing relative to the EEG onset and clinical onset, which are annotated by medical experts. We define the period $< 2 min$ after clinical onset as the ictal period, and the period $< 30 min$ before EEG onset as the interictal period.  Given that patients’ behaviors often remain unchanged for extended periods, we strive to include a variety of actions in the extracted clips. The transition period usually ranges from $1 s \sim 20 s$ after EEG onset for each seizure event. In our dataset, the average transition period is approximately $18.2 s$ across 33 seizures.
To address the imbalance in duration between interictal and other periods, we extract clips from interictal period without overlapping, and from both transition and ictal periods with $4 s$ overlappings. For fair performance comparison, we randomly extract $20\%$ clips from both transition and ictal periods from each seizure, and an equivalent number of clips (20\% transition + 20\% ictal) are extracted from the interictal period for validation, the rest clips are for model training. In this study, we aim to train a generalized model applicable to all patients rather than a patient-specific model.

\section*{B. Detailed architecture of VSViG model}
In \Cref{tab:archi}, we present detailed architecture of proposed VSViG model and its lighten version.

\begin{table}[ht]
    \caption{\textbf{Details of two versions of VSViG.} In Stem layer, $32\times 32, C$ is kernel size and output channels of Conv2D implemented on input patches ($H\times W \time 3$). Then we utilize 4-stage spatiotemporal modeling to process features. $E, C$ denotes the number of constructed edges between neighbors and output channels for spatial modeling, $1\times K_{S} \times K_{T}, C$ denotes kernel size and output channels of 3D-CNN in temporal modeling.}
  \centering
  
  \renewcommand{\arraystretch}{1.05}
  \small
  \begin{tabular}{c|c|c}
  \hline
  \makecell{Layer \\ name}& \makecell{Output \\size T} & VSViG-Light/VSViG\\
  \hline
  (Patches)&30&  $  (1\times 15\times T \times 32 \times 32  \times 3)$ \\
  \hline
  Stem & 30& $32\times32,12/24$ \\
  \hline
  (Features)&30&  $  (1\times 15\times T \times C)$ \\
  \hline 
  
  Stage1 & 30 
  & $\begin{bmatrix} E=12,12/24\\ E=2,12/24\\ 1\times1^{2},12/24\\  1\times3^{2},12/24\\  1\times1^{2},24/48 \end{bmatrix}\times2$ \\
  \hline 
  Stage2 & 15
  & $\begin{bmatrix}  \ E=12,24/48\\ \ E=2,24/48\\ \ 1\times1^{2},24/48\\  \ 1\times3^{2},24/48\\  \ 1\times1^{2},48/96 \end{bmatrix}\times2$ \\
  \hline
  Stage3 & 8
  & $\begin{bmatrix}  \ E=12,48/96 \\ \ E=2,48/96\\ \ 1\times1^{2},48/96\\  \ 1\times3^{2},48/96\\  \ 1\times1^{2},96/192 \end{bmatrix}\times6$ \\
  \hline
  Stage4 & 4
  & $\begin{bmatrix}  \ E=12,96/192 \\ \ E=2,96/192\\ \ 1\times1^{2},96/192\\  \ 1\times3^{2},96/192\\  \ 1\times1^{2},192/384 \end{bmatrix}\times2$ \\
  \hline
  Head & 1& GAP, 1-d FC, Sigmoid \\
  \hline
  \multicolumn{2}{c|}{FLOPS}& 0.44G/1.76G\\

  \hline
  \end{tabular}

  \label{tab:archi}
\end{table}

\section*{C. Detailed seizure detection performance}
In \Cref{tab:detailed-performance}, detailed patient information and seizure detection performance per patient are shown.

\begin{table*}[t]
\centering
\footnotesize
  \caption{\textbf{Detailed seizure detection performance per patient.} M: Male, F: Female, Sen.: Sensitivity (\# detected seizure / \# total seizure), FDR: False Detection Rate, DT: Decision Threshold, AP: Accumulative Probability strategy, P: Partial (focal) seizures, PG: Partial to Generalized tonic-clonic seizures.}
\renewcommand{\arraystretch}{1.1}
  \begin{tabular}{c c c c c c c |c c |c c}
  \specialrule{.1em}{.05em}{.05em}
   \multirow{2}{*}{PatID}  & \multirow{2}{*}{Age} & \multirow{2}{*}{Sex} & \multirow{2}{*}{Type} & \multirow{2}{*}{\# sei.} & \multirow{2}{*}{Sen.} & \multirow{2}{*}{FDR}  & \multicolumn{2}{c|}{$DT$= 0.3 w/o AP} & \multicolumn{2}{c}{$DT$= 0.3 w. AP}\\
   &&&&&&&$L_{EO}$&$L_{CO}$&$L_{EO}$&$L_{CO}$ \\
  \specialrule{.1em}{.05em}{.05em}
  pat01& 28& M & PG&2&2/2 &0 & 11$s$ &-2.2$s$& 1.3$s$ & -12$s$ \\
  pat02& 37& M& PG&2&2/2 &0&  41.7$s$ & -8.5$s$ & 20.7$s$ &-31.5$s$ \\
  pat03& 39& M& PG& 2&2/2 &0& 4$s$ & -2.2$s$ & 3$s$ & -3.2$s$ \\
  pat04& 33& F& P& 1&1/1 &0 & 23.5$s$ & -8$s$  & 12.5$s$ & -19$s$ \\
  pat05& 35& F& PG& 3&3/3 &0& 16$s$ & -4.1$s$  & 5.1$s$ & -15$s$ \\
  pat06& 30& M& P& 3&3/3 &0& 1.7$s$ & -1$s$ &  1.7$s$ & -1$s$ \\
  pat07& 32& M& PG& 2&2/2 &0& 32$s$ & -8$s$ & 8$s$ & -32$s$ \\
  pat08& 57& F& P& 3&3/3 &0&  6$s$ & -2.6$s$ & 2$s$ & -6.6$s$ \\
  pat09& 31& M& P& 3&3/3 &0&  6.1$s$ & -1.5$s$  & 5.5$s$ & -2.1$s$ \\
  pat10& 24& F& P& 2&2/2 &0&  3.7$s$ & -1.5$s$ & 3.2$s$ & -1.9$s$ \\
  pat11& 55& F& P& 3&3/3 &0& 11$s$ & -3.1$s$ & 5$s$ & -8.1$s$ \\
  pat12& 23& M& P& 4&4/4 &0&  4.2$s$ & -1.9$s$  & 4$s$ & -2.1$s$ \\
  pat13& 29& F& PG& 1&1/1 &0 & 0.5$s$ & -5$s$ & 0.5$s$ & -5$s$ \\
  pat14& 18& M& PG& 2&2/2 &0& 49.2$s$ & -17.5$s$  & 16$s$ & -50.7$s$ \\
  \specialrule{.1em}{.05em}{.05em}

  \end{tabular}

  \label{tab:detailed-performance}
\end{table*}

\end{document}